# Predicting the performance of hybrid ventilation in buildings using a multivariate attention-based biLSTM Encoder – Decoder

**Gaurav Chaudhary[1*], Hicham Johra[2], Laurent Georges[1], Bjørn Austbø[1]**

[1]Department of Energy and Process Engineering & ENER-SENSE, NTNU, Trondheim, Norway
[2]Department of the Built Environment, Aalborg University, Aalborg Øst, Denmark
[*]Corresponding author: gaurav.chaudhary@ntnu.no

**Abstract**. Hybrid ventilation is an energy-efficient solution to provide fresh air for most climates, given that it has a reliable control system. To operate such systems optimally, a high-fidelity control-oriented modesl is required. It should enable near-real time forecast of the indoor air temperature based on operational conditions such as window opening and HVAC operating schedules. However, physics-based control-oriented models (i.e., white-box models) are labour-intensive and computationally expensive. Alternatively, black-box models based on artificial neural networks can be trained to be good estimators for building dynamics. This paper investigates the capabilities of a deep neural network (DNN), which is a multivariate multi-head attention-based long short-term memory (LSTM) encoder-decoder neural network, to predict indoor air temperature when windows are opened or closed. Training and test data are generated from a detailed multi-zone office building model (EnergyPlus). Pseudo-random signals are used for the indoor air temperature setpoints and window opening instances. The results indicate that the DNN is able to accurately predict the indoor air temperature of five zones whenever windows are opened or closed. The prediction error plateaus after the 24$^{th}$ step ahead prediction (6 hr ahead prediction).

## 1. Introduction
Buildings are responsible for over 40% of global energy use and 36% of greenhouse gas emissions, with heating, ventilation, and air conditioning (HVAC) operation accounting for almost half of it [1]. Reducing the overall building energy demand and footprint has thus become an urgent task to meet the current sustainability goals and tackle current energy crises. To that end, natural ventilation is seen as one of the most effective passive energy-saving measures for buildings [2]. HVAC systems in buildings coupled with natural ventilation (hybrid ventilation) can theoretically provide the most energy efficient system for any climate, given that they have a fast and reliable control system. For such systems, a high-fidelity control-oriented prediction model is required. It should be able to forecast building dynamics in near-real time for given operational conditions such as window opening and HVAC operational schedules. This is, however, challenging due to the time-varying building dynamics, disturbances from occupants, lighting and plug-in loads, and external factors like outdoor weather. Developing efficient prediction models accounting for these building dynamics has been a bottleneck to implementing predictive-based control strategies [3].

Buildings with hybrid ventilation is particularly challenging for dynamic modelling. When natural ventilation occurs, e.g., when opening a window, the indoor temperature variation depends on many parameters, such as the indoor-outdoor temperature difference, the window opening configuration and effective opening area, the HVAC mode, and the internal loads. Such building dynamics can be fully modelled using well-established laws of physics (i.e., white box approach) [3] or these laws provide the model structure while meausurement data is used to calibrate the model parameters (i.e., grey box approach) [4]. However, a typical white box modelling tool like EnergyPlus or IDA-ICE requires expert efforts to define, set and adjust the multiple model parameters. A grey box model, such as a resistor-capacitor (RC) network, requires a robust estimation of its parameters.

Reinforced by the massive amount of data induced by the deployment of metering and sensing technologies in buildings, the data-driven black box approach for building dynamics prediction has increased in popularity in recent years [5]. Black box models can have the advantage of low development costs and scalability. However, such models usually require a large amount of training data to perform adequately. This, however, can be solved by using transfer learning methods that couple data generated from white box modelling tools [6] and system identification techniques. A black box model pre-trained on various operating conditions and scenarios simulated with the white box model of a generic building could, in theory, be suitable for real building applications after only tuning the former with a very small dataset [6–8].

Following that principle, it is hypothesized that deep neural networks (DNNs) can be employed as accurate black box models for the prediction of indoor environment. DNNs are suitable for complex building dynamics as they can handle non-linear multivariable modelling situations. It was shown that a neural network with enough hidden layers can approximate arbitrary continuous functions defined on a closed and bounded set [9]. DNNs based on convolutional neural networks (CNNs) and recurrent neural networks (RNNs) like long short-term memory (LSTM) [10] and gated recurrent units (GRUs) [11] have been widely used in applications like speech recognition [12], natural language processing [13] and computer vision [14]. Like building energy models, these applications use data in the form of time series. In RNNs such as LSTMs or GRUs, each input corresponds to an output for the same time step. However, in many real cases, there is a need to predict an output sequence given an input sequence of different lengths without correspondence between each input and output. This situation is called sequence-to-sequence Mapping (also known as Encoder-Decoder models)and lies behind commonly used applications like machine translation, question answering, chat question answering, chat-bots, and text summarization. The most commonly used neural network unit in Encoder-Decoder models is the LSTM unit. However, they seem to suffer from short-term memory over long time series sequences. Advancements like the attention mechanism [15] and the transformer [16] used in conjunction with RNNs have improved their prediction performance [17]. The attention mechanism improve the model's accuracy by giving higher weights to relevant parts of the sequence and vice versa for irrelevant parts [18]. The multi-head attention (MHA) module introduced in the transformer model [16] runs through the attention mechanism several times in parallel, attending to different parts of the sequence differently. Compared to LSTMs, MHA retains direct connections to all previous timestamps in the sequence, allowing information to propagate over much longer sequences. To summarize, the RNNs are excellent at capturing the local temporal characteristics of a sequence, while the transformer model can learn long-term dynamics.

Standard DNNs are deterministic in nature and always, in theory, produce uncertain results due to model uncertainity and data uncertainty. Model uncertainty accounts for uncertainty in tunable parameters in a model whereas data uncertainty accounts for noisy and out-of-distribution data. Probabilistic DNNs account for such uncertainty in the final results by producing prediction intervals. In line with other time series forecasting models [19], the DNN model developed for this study also generates prediction intervals on top of point forecasts. This is done by simultaneous prediction of various percentiles ($50^{th}$, $90^{th}$, $95^{th}$ and $99^{th}$) at each time step using quantile regression [20].

This paper investigates if and how a DNN can be trained to predict the effect of the window opening on the indoor air temperature dynamics in a conditioned building. The training data for this DNN was generated using the white box building energy simulation tool EnergyPlus.

## 2. Model architecture

The encoder–decoder model developed for this study employs both RNNs, specifically bi-directional LSTMs (biLSTM) [21], and several components of the transformer model such as Self-MHA, Cross-MHA and Gated Residual Networks (GRNs) [22]. A biLSTM is a sequence processing model that consists of two LSTMs: one taking the input in a forward direction (past to future) and the other in a backwards direction (future to past).). This effectively improves the contextual information of the data dynamics. The Self-MHA components in both the Input Encoder and the Input Decoder are used to determine long-term relationships within input data, producing attention scores for the biLSTMs. The Cross-MHA takes the representation of both the encoder input sequence and decoder input sequence coming from biLSTMs and learns the relationships for larger periods producing an attention score for Output Decoder biLSTM. This improves the context for short-term dependencies. Residual connection [23] in the form of GRN [22] is applied over each module by first applying component gating layers based on Gated Linear Units (GLUs) [24], followed by layer normalization [25]. GLUs allow the model to control the extent to which the residual connection mechanism contributes to the original input. In this study, the model takes the past seven days of data as input and predicts 24 hours into the future. The data has a temporal granularity of 15 minutes, so the model takes 672 data points from the past and predicts 96 data points. The model takes two sets of inputs: "*Known past inputs*" for the input encoder and *"Known future inputs"* for the input decoder. "*Known past inputs*" are the past seven days of weather data, time information, and zone-specific information like occupancy, external loads, temperature setpoints in the zones, and actual indoor air temperatures (IATs) of these zones. The input data also includes the opening sensor signals of the windows. *"Known future inputs"* are the future 24 hours of weather forecast, the control variables which are the heating setpoints of HVAC systems, and the window opening signals. These inputs are similar to what a zone-level controller can access to condition a zone. Here, the weather forecast is created from the actual weather data, with a added gaussian noise of zero mean and standard deviation of 0.01 °C. The model's output is the future 24 hours of IATs for all zones. A schematic of the model structure is shown in Figure 1.

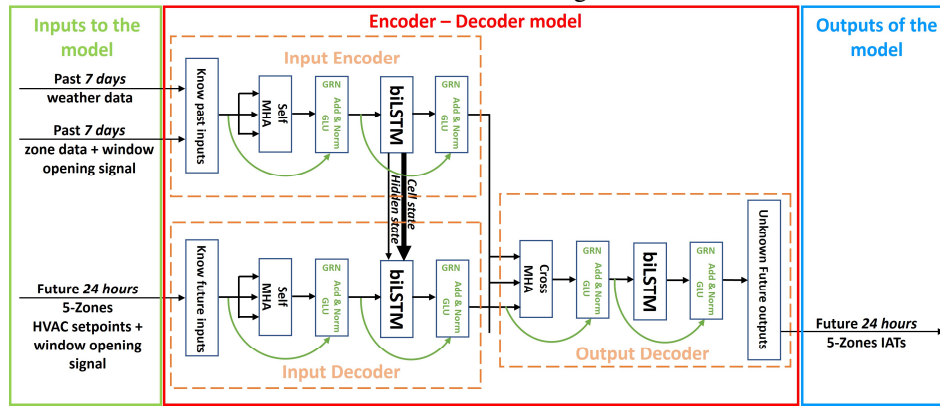

**Figure 1.** Encoder-decoder model for predicting building dynamics using both recurrent networks and components of the transformer model.

## 3. Study case description and data generation

Training data was generated from a small-size office building modelled with EnergyPlus v22.1.0. The building model is a generic 5-zone EnergyPlus example file geometry. The building is a single-floor rectangle of dimensions 30 m x 15 m , with a ceiling height of 2.4 m. It has four exterior zones and one interior zone (see Figure 2). There are windows on all four facades, and glass doors on the south-west and north-east facades. Overhangs shade the south-facing window and door. There is no internal opening between the zones. The U-values of the internal and external walls are 1.6 W/m²K and 2.8 W/m²K, respectively. All fenestrations are high-performance windows with a U-value of 0.7 W/m²K. To reduce overheating, automatic window shading control lowers the interior shade when the outside temperature exceeds 23 °C. The building uses a variable refrigerant flow HVAC system for conditioning the zones, whereas the ventilated air for the building is delivered by a dedicated outdoor air system.

The schedules for occupancy and miscellaneous electric loads are generated by an agent-based stochastic occupancy simulator [26]. The lighting schedule is based on standard working hours from 07:00 to 19:00.

To "excite" the DNN for all possible changes in heating setpoints during training (i.e., create sufficient variability in key input variables of the training dataset), a multi-pseudo random sequence (m-PRS) input signal is applied to the temperature heating setpoints of the five zones. During occupied hours, the m-PRS signals change randomly between 18 and 22 °C (with 0.5 °C intervals) and stay at that value for a random amount of time. The cooling setpoint is 5 °C above the heating setpoint. For non-occupied hours, the heating and cooling setpoints drop to a setback of 15 and 30 °C, respectively. The signal excitation method is a system identification method aiming to excite one input with a signal that is not correlated with other inputs. The DNN thus learns the underlying dynamics of thermal setpoints [27].

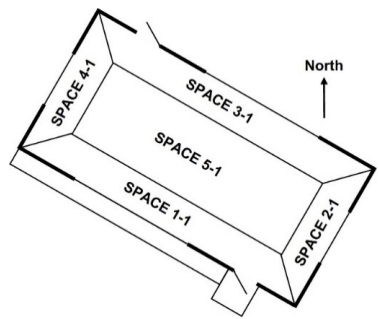

**Figure 2.** 5-zone office building simulated in EnergyPlus for generating dataset.

The windows' opening/closing is modelled with the *ZoneVentilation:WindandStackOpenArea* EnergyPlus object. It allows to define the limits on the outdoor conditions (temperature, wind speed) that determine whether the window is open or closed. The equation used to calculate the wind-driven natural ventilation rate is based on the "*Wind and Stack with Open Area*" model. Pseudo-random binary sequence (PRBS) signals are used to actuate the opening and closing of the windows to excite the DNN. In the PRBS signal a random 1 represents that a window is opened for the next 30 minutes. Figure 3 shows the indoor air temperature of Space 1-1 along with the mPRS input signal for heating and cooling setpoints, the outdoor air temperature and the PRBS window opening/closing signal. The effect of opening windows can be observed as a sharp decline followed by a gradual rise of the IAT.

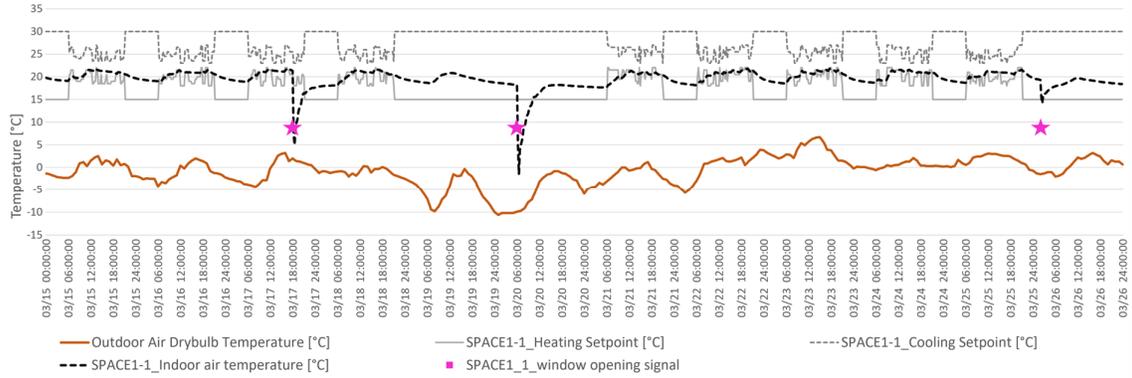

**Figure 3.** The indoor air temperature of Space 1-1 along with mPRS input signal for heating and cooling setpoints, outdoor air temperature and opening/closing signal.

**Table 1.** Multivariate inputs-output variables used for the study

| Input variables (*Known past inputs*) | | | Interval | Input variables (*Known future inputs*) | | | Interval |
|---|---|---|---|---|---|---|---|
| $T^{out}$ | Outside Dry-bulb temperature | °C | [ -30.0, 40.0] | $T^{out}$ | Forecasted Outside Dry-bulb temperature | °C | [ -30.0, 40.0] |
| $H^{out}$ | Relative humidity of air | % | [ 0, 100] | $H^{out}$ | Forecasted Relative humidity of air | % | [ 0, 100] |
| $W^{out}$ | Wind speed | m/s | [ 0, 25] | $W^{out}$ | Forecasted Wind speed | m/s | [ 0, 25] |
| $I\_Norm^{out}$ | Direct normal radiation | W/m$^2$ | [ 0, 1300.0] | $I\_Norm^{out}$ | Forecasted Direct normal radiation | W/m$^2$ | [ 0, 1300.0] |
| $I\_Hor^{out}$ | Diffuse radiation on horizontal surface | W/m$^2$ | [ 0, 1300.0] | $I\_Hor^{out}$ | Forecasted Diffuse radiation on horizontal surface | W/m$^2$ | [ 0, 1300.0] |
| $h$ | Sine and Cosine of the Hour of the day | - | [ -1.0, 1.0] | $h$ | Sine and Cosine of the Hour of the day | - | [ -1.0, 1.0] |
| $d$ | Sine and Cosine of the Day of the week | - | [ -1.0, 1.0] | $d$ | Sine and Cosine of the Day of the week | - | [ -1.0, 1.0] |
| $m$ | Sine and Cosine of the Month of the year | - | [ -1.0, 1.0] | $m$ | Sine and Cosine of the Month of the year | - | [ -1.0, 1.0] |
| $hol$ | Holiday | Boolean | [ 0, 1] | $hol$ | Holiday | Boolean | [ 0, 1] |
| $E_i$ | Equipment load of the zone $i(1-5)$ | W | [ 0.0, 1000.0] | $WS_i$ | Window opening signal of zone window $i(1-4)$ | Boolean | [ 0, 1] |
| $Occu_i$ | Occupancy in the zone $i(1-5)$ | - | [ 0, 30] | $SP_i$ | Heating Setpoint of zone $i(1-5)$ | °C | [ 15.0, 30.0] |
| $WS_i$ | Window opening signal of zone window $i(1-4)$ | Boolean | [ 0, 1] | **Output variables** (*Unknown future outputs*) = $y_{i(t, t+96)}$ | | | |
| $SP_i$ | Heating Setpoint of zone $i(1-5)$ | °C | [ 15.0, 30.0] | | | | |
| $T^{in}_i$ | Indoor temperature of zone $i(1-5)$ | °C | [ 10, 40.0] | $T^{in}_i$ | Indoor temperature of zone $i$ | °C | [ 10, 40.0] |

To predict the IAT, essential features that could be commonly available in an office building management system are selected as inputs (see Table 1). In contrast, zone-specific features like equipment and occupancy can be deduced from $CO_2$ concentration monitoring [28]. The hour of the day, the day of the week, the month of the year, and the holiday schedule were also used as inputs. The hour input leads to knowing the difference between the temperature profile during the occupied and unoccupied time and understanding the daily dynamics of the building. Day and holiday input leads to distinguishing between business and weekend days.

## 4. Training procedure

The 12 months of data is split into training (60 %), validation (20 %) and testing (20 %). Deep learning models perform better when numerical input variables are scaled to a standard range. For this study all the input variables in the dataset were scaled to the range [-1.0, 1.0] using *MinMaxScaler*. The minimum and maximum value of the scaling is mentioned in Table 1 in the interval column. The DNN is implemented with *Tensorflow 2.11.0* and the *Keras* library in *Python 3.10.0*. The hyperparameters of the model and training are listed in Table 2.

**Table 2.** Hyperparameters for the deep neural network

| Hyperparameter name | Description | Value |
|---|---|---|
| n_past | Number of past data points for input data | 672 |
| n_future | Number of future data points predicted | 96 |
| noise_sd | Standard deviation for Gaussian noise for weather data used in *Known future input* | 0.01 |
| batch_size | Diffuse radiation on horizontal surface | 256 |
| optimizer | Loss optimizer used for training | ADAM |
| loss | Loss function for the model | quantile loss |
| mha_head | Number of multi-head attention module | 4 |
| dropout | Dropout % used for training | 0.3 |
| RNN | Type of recurrent neural network units used | biLSTM |
| RNN - units | Number of recurrent neural network units whenever used | 200 |
| Activation | Activation function used in RNN units | tanh |

The coefficient of variance of the root mean squared error (CVRMSE) is used as the accuracy evaluation metric to compare the predicted IATs and the actual ones.

## 5. Results

For the testing, 96 steps, i.e. 24 hr ahead, prediction was done for 5112 instances. Figure 4 shows 3 selected instances for two of the zones (SPACE1-1 and SPACE2-1). Each subplot shows the actual IAT, heating setpoint and window opening signal (presented as a pink start marker) and probabilistic forecast of IAT for that instance. Each forecast has 90%, 95% and 99% confidence intervals presented in decreasing opacities of green, and the 50$^{th}$ quantile in red. The various instances of prediction are selected to have a mix of both bad and good IAT forecasts, window opening signals both few and many steps ahead, and window opening signals with low and high influence on the IAT. While Figure 4 shows the instances of prediction for the whole 96 steps, Figure 5 focuses on the 24$^{th}$ step (six hours) ahead for all instances of the same two zones. The results for other zones can be seen in the supplementary material.

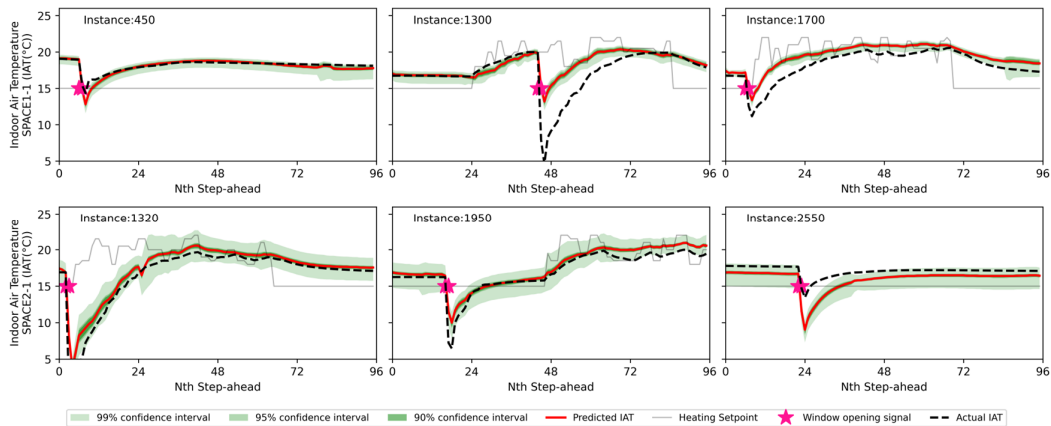

**Figure 4.** The real vs predicted indoor air temperature of two zones at various instances of prediction. Extended high resolution results: https://github.com/gaurav306/NSB23-Predicting-the-performance-of-hybrid-ventilation-in-buildings-using-a-multivariate-attention-/blob/main/Figure-4.png

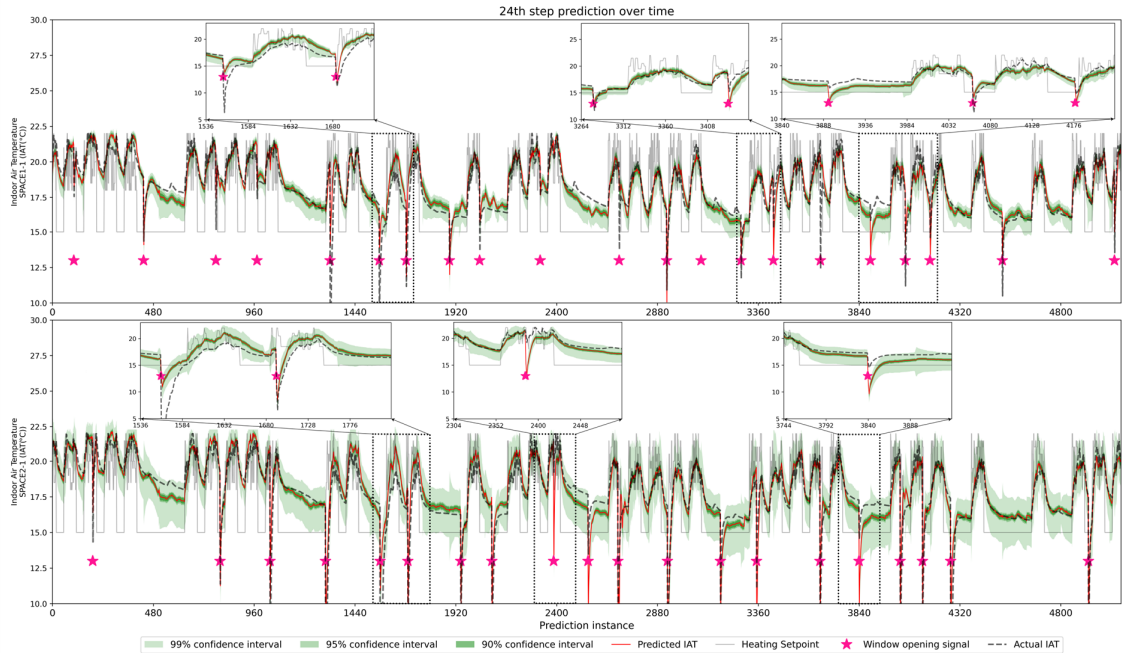

**Figure 5.** The real indoor air temperature vs 24$^{th}$ step prediction done at all instances of prediction for SPACE1-1 and SPACE2-1. Some of the bad predictions when the window is opened are shown in various zoomed subplots. The high-resolution version of the figure above along with wider figures for 1$^{st}$, 24$^{th}$, 48$^{th}$, 72$^{nd}$ and 96$^{th}$ step prediction over time, is available at: https://github.com/gaurav306/NSB23-Predicting-the-performance-of-hybrid-ventilation-in-buildings-using-a-multivariate-attention-/blob/main/Figure-5.pdf

These qualitative results show that a DNN can predict well the IAT with a random heating setpoint signal and a sharp decrease and gradual rise of temperature when a window is opened. In some instances where the error is high, like Instance:1300 of SPACE1-1 and Instance:2550 of SPACE2-1, the prediction is good when the control signal is closer to t=0 step ahead. As observed in Figure 4, window opening severely affected the IAT of SPACE2-1. This may be due the wind predominantly flowing from southeast to northwest for the testing period, adding more randomness to the indoor temperature dynamics. This could potentially be solved by adding wind direction to the input data.

The quantitative error in the predictions is given for all zones in Figure 6. It shows how the error has changed for certain steps into the future for all instances. The CVRMSE (%) shown in the plot is the error between the actual IAT and the 50$^{th}$ quantile of the predicted IAT. The error for all the zones plateaus after the 24$^{th}$ step ahead, which can be a good sign of prediction stability. However, the error of all zones is higher at the initial steps, which can be due to sudden inaccuracies of forecasted weather used as input data.

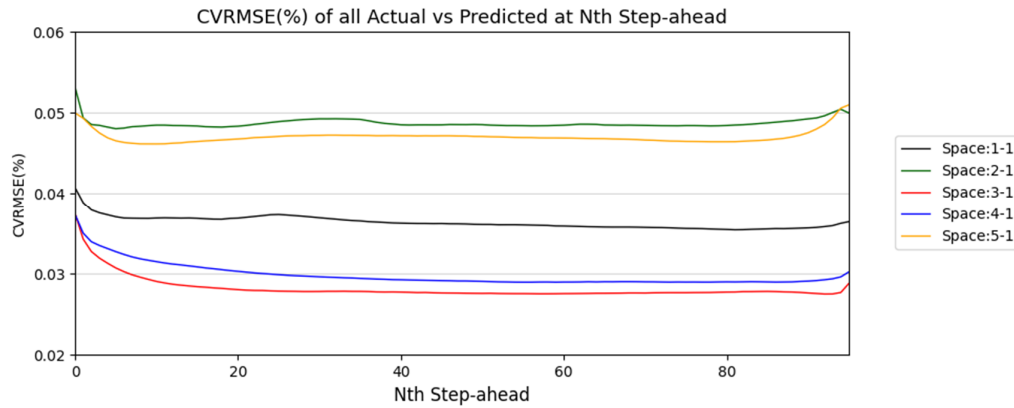

**Figure 6.** The forecast error of actual temperature vs 50[th] quantile predicted for all N[th] step-ahead.

## 6. Discussion and conclusions

The DNN model presented in this paper is fairly advanced in comparison to other time series prediction models and sequence-to-sequence neural networks commonly found in the literature about building energy simulation. Other statistical time series models, such as ARX ARMAX, are linear and time-variant in nature. However, they perform poorly when presented with nonlinearities and sudden uncertainties in the system. Through tests it was seen that sequence-to-sequence neural networks using RNNs in their basic form, although nonlinear in nature, are not able to capture varying building dynamics. Both the above-mentioned models are also not able to take input such as heating setpoints and window opening signals separately. The complexity of the final model structure presented in the current article is the result of many model iterations developed after analyzing the limitations of other models, and the will to improve control-oriented model building applications. Using transformer model components enables to capture long-term building dynamics with high accuracy.

This paper indicates that a deep learning-based neural network can be used as an estimator of building IAT given a heating setpoint and control signal for the window opening. It was observed that the prediction model can predict the incoming drop in IAT when a window is expected to be open. This can reversed: window opening can be actuated based on the model's predictions to accurately regulate the indoor thermal comfort with minimum energy use. The training of this DNN can be extended to predict other indoor comfort criteria-related features like the relative humidity or the $CO_2$ concentration, as well as energy demand. Such models can also be employed in model predictive control [29] or a reinforcement learning architecture where all controllable building system parameters are optimized to maximize indoor comfort and minimize the costcosts based on a penalty signal (e.g.,emissions, energy use).

This paper also showed that the transfer learning approach can be a very effective method to train DNN models for various building energy and environment applications. The prediction model was tested on boundary conditions, occupancy schedules, heating setpoints and window opening signals that were not seen before during the training phase. This transfer learning approach for system identification can be extended where merely a simple representative energy model of a building is required.

**Acknowledgements**
The authors acknowledge the support from the strategic research program ENERSENSE at the Norwegian University of Science and Technology (NTNU).